# Extend natural neighbor: a novel classification method with self-adaptive neighborhood parameters in different stages

Ji Feng[a], Qingsheng Zhu[a,1], Jinlong Huang[a], Lijun Yang[a]

[a]Chongqing Key Lab. of Software Theory and Technology, College of Computer Science, Chongqing University, Chongqing 400044, China

**Abstract**

Various kinds of *k*-nearest neighbor (KNN) based classification methods are the bases of many well-established and high-performance pattern-recognition techniques, but both of them are vulnerable to their parameter choice. Essentially, the challenge is to detect the neighborhood of various data sets, while utterly ignorant of the data characteristic. This article introduces a new supervised classification method: the extend natural neighbor (ENaN) method, and shows that it provides a better classification result without choosing the neighborhood parameter artificially. Unlike the original KNN based method which needs a prior *k*, the ENaNE method predicts different *k* in different stages. Therefore, the ENaNE method is able to learn more from flexible neighbor information both in training stage and testing stage, and provide a better classification result.

## 1. Introduction

With the continuous expansion of data availability in many areas of engineering and science, it becomes critical to identify patterns from vast amounts of data, to identify members of a predefined class, which is called as classification. Therefore, classification becomes a fundamental problem, especially in pattern recognition and data mining, and several effective algorithms have been successfully applied in many real-world applications. In the setting of a classification, there are two kinds of classifiers, parametric classifiers and nonparametric classifiers. With the coming of the Big Data era [1] [2], nonparametric classifiers have received particular attention because the data distributions of many classification problems are either unknown or very difficult to obtain in practice. *K*-nearest neighbor (KNN) classifiers, as one of the typical representative of nonparametric classifiers, is a basic task which is used to classify a query object into the category as its most nearest example [3] [4]. The *k*-nearest neighbor (KNN) classifier extends this idea by taking *k* nearest points and assigning the sign of the majority.

[1] Corresponding author: Tel.: +86-023-65105660; fax: +86-023-65104570 ;
*Email address:* qszhu@cqu.edu.cn (Qingsheng Zhu)

KNN classifier has attracted many researchers to make efforts [5], and is applied in various domains [6-8]. Therefore, it has been witnessed in considerable applications in many different disciplines [9-14]. However, there are still following two main problems to limit the usage of KNN classifier:

The efficiency of KNN classification heavily depends on the type of distance measure, especially in a large-scale and high-dimensional database. In some applications, the data structure is so complicated that the corresponding distance measure is computationally expensive [15, 16]. KNN classifier has to compare it with all the other examples in the database for each query. It may become impractical for a huge database and frequent queries.

Traditional *k*-NN adopts a fixed *k* for all query samples regardless of their geometric location and related specialties. Furthermore, those *k*-nearest neighbors may not distribute symmetrically around the query sample if the neighborhood in the training set is not spatially homogeneous. The geometrical placement might be more important than the actual distance to depict a query samples neighborhood.

Therefore, to improve neighborhood based classifiers, more algorithms are proposed. Recently, based on mutual *k*-nearest neighborhood (M*k*NN) method, Tang and He propose an Extended nearest neighbor (ENN) method for classification which makes use of the two-way communication style [17]. Unlike the classic KNN rule which only considers the nearest neighbors of a test sample to make a classification decision, ENN method considers not only who the nearest neighbors of the test sample are, but also who consider the test sample as their nearest neighbors.

This two-way communication style is an advantage and also a disadvantage of ENN method. The advantage means that the classification decision of ENN method is depended by all variable training data. The disadvantage means that the problem of parameter selection such as the selection of *k* in KNN still exists. ENN has a parameter of *k*, based on which, the size and shape of the graph can be changed. This makes the further optimization possible; however, there exists the problem of parameter selection.

To overcome the above limitations of ENN classification, in this work, we propose a new method called as ENaN method, which keep the advantage and solve the disadvantage of ENN method, with the help of natural neighbor (NaN) method to choosing the optimum value of *k* both in training stage and testing stage. The rest of the paper is outlined as follows: In the next section, we will survey related works. We will present our proposed method in detail in Section 3, followed by a series of experiments in Section 4. Section 5 concludes this paper with discussions on future works.

2. **Background and notation**

   *2.1. ENN method*

ENN method makes a prediction in a two-way communication style: it considers not only who the nearest neighbors of the test sample are, but also who consider the test sample as their nearest neighbors. By exploiting the generalized class-wise statistics from all training data by iteratively assuming all the possible class memberships of a test sample, the ENN is able to learn from the global distribution, therefore improving pattern recognition performance and providing a powerful technique for a wide range of data analysis applications.

The generalized class-wise statistics of ENN method with two-class classification problem is defined as follow:

**Definition 1 (generalized class-wise statistics).** Generalized class-wise statistics $T_i$ for class *i* is defined as follow:

$$T_i = \frac{1}{n_i k} \sum_{x \in S_i} \sum_{r=1}^{k} I_r(x, S = S_1 \bigcup S_2) \qquad i = 1, 2 \qquad (1)$$

Where $S_i$ denote the samples in class *i*, respectively, *x* denotes one single sample in $S = S_1 \cup S_2$, $n_i$ is the number of samples in $S_i$, and *k* is the user defined parameter of the number of the nearest neighbors. $I_r(x, S)$ indicates whether both the sample *x* and its r-th nearest neighbor belong to the same class, defined as follows:

$$I_r(x, S) = \begin{cases} 1, & if\, x \in S_i\, and\, NN_r(x, S) \in X_i \\ 0, & otherwise \end{cases} \qquad (2)$$

## 2.2. NaN method

Recently, our team has presented a new parameter-free nearest neighbor method called Natural Neighbor (NaN) [18]. NaN method is inspired by the friendship of human society and could be regarded as belonging to the category of scale free nearest neighbor method, and it's an effective method in outlier detection [19]. Natural Neighbor method makes three key contributions to the current state:

1. Natural Neighbor method can create an applicable neighborhood graph based on the local characteristics of various data sets. This neighborhood graph can identify the basic clusters in the data set, especially manifold clusters and noises.

2. This method can provide a numeric result named Natural Neighbor Eigenvalue (NaNE) to replace the parameter k in traditional KNN method, and the number of NaNE is dynamically chosen for different data sets.

3. The natural neighbor number of each point is flexible, and this value is a dynamic number ranging from 0 to NaNE. The center point of the cluster has more neighbors, and the neighbor number of noise is equal to 0.

NaN method is defined as follow:

**Definition 2 (Natural Neighbor Eigenvalue).** When the algorithm reaching the Stable Searching State, Natural Neighbor Eigenvalue (NaNE) is equal to the searching round r.

$$\lambda \triangleq r_{r \in N}\{r | (\forall x_i)(\exists x_j)(r \in N) \wedge (x_i \neq x_j) \rightarrow (x_i \in KNN_r(x_j)) \wedge (x_j \in KNN_r(x_i))\} \quad (3)$$

**Definition 3 (Natural Neighbor).** Natural neighbor of xi is defined as follow

$$x_j \in NN(x_i) \Leftrightarrow (x_i \in KNN_\lambda(x_j)) \wedge (x_j \in KNN_\lambda(x_i)) \quad (4)$$

3. **Proposed approach: ENaN method for classification**

*3.1. combined points of NaN and ENN*

Original ENN classification has two stages: training stage and testing stage. Training stage can be developed to calculate the generalized class-wise statistic $T_i$ for each class to build weighted KNN graphs, in which training samples are the vertices and the distances of the sample to its nearest neighbors are the edges. In testing stage, firstly we assume test sample belongs to a class and calculate the generalized class-wise statistic $T_i$ for each class. Then we assume it belongs to another class and calculate $T_i$ for each class. After all class assumptions are done, a classification decision is made according to Eq.1.

In original ENN classification, neighborhood parameter *k* is used to calculate the generalized class-wise statistic $T_i$ according to Eq.1. Therefore, similar to traditional KNN methods, the classification result of ENN methods depends heavily on the selection of the neighborhood parameter *k*. The classification accuracies of KNN and ENN methods over a range of *k* is illustrate in Fig.1.

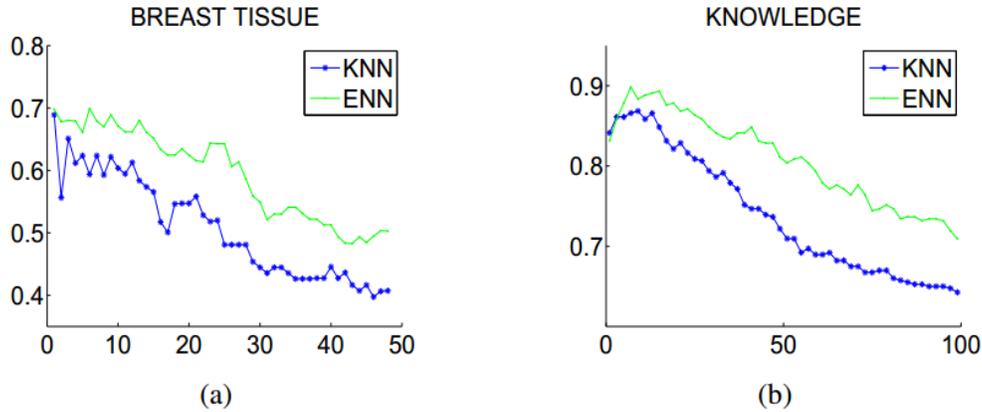

Figure 1: Classification accuracies of KNN and ENN methods over a range of k. The data sets are standard UCI data sets

Fig.1 firstly demonstrate the effectiveness of ENN method. The Classification accuracies of ENN method are significantly better than KNN method in most cases. Further more, it is clear that the parameter choice problem is still exist in ENN, a terrible selection of *k* will also shapely decrease the accuracy of classification.

NaNE method can help ENN based method choose the parameter *k* without any priority knowledge. Considering about the specific situations of ENN method's two stages, our NaN method selects neighborhood parameters with different ways. In training stage, ENN method needs a neighborhood to measure the distribution of training data set, in this case NaNE which reflects the overall distribution of the data is the better choice. To illustrate the relationship between data set and its natural neighbor eigenvalue, four simple examples of artificial data sets are shown in Fig.2 .

But when the unknown sample comes in testing stage, a self-adaptive neighborhood of each sample will be more accurate. Fortunately, in the idea of natural neighbors, the number of neighbors at each point is not fixed. Natural neighborhood search process does not need the limitation of neighborhood parameters, and this dynamic neighborhood value can more accurately reflect the relationship between data points, so that each data point according to their own environmental characteristics to find a suitable neighbor. In particular, data with more neighbors is denser, while edge points have fewer natural neighbors. Fig.3 shows dynamic neighborhood value of natural neighbor method in two UCI data sets.

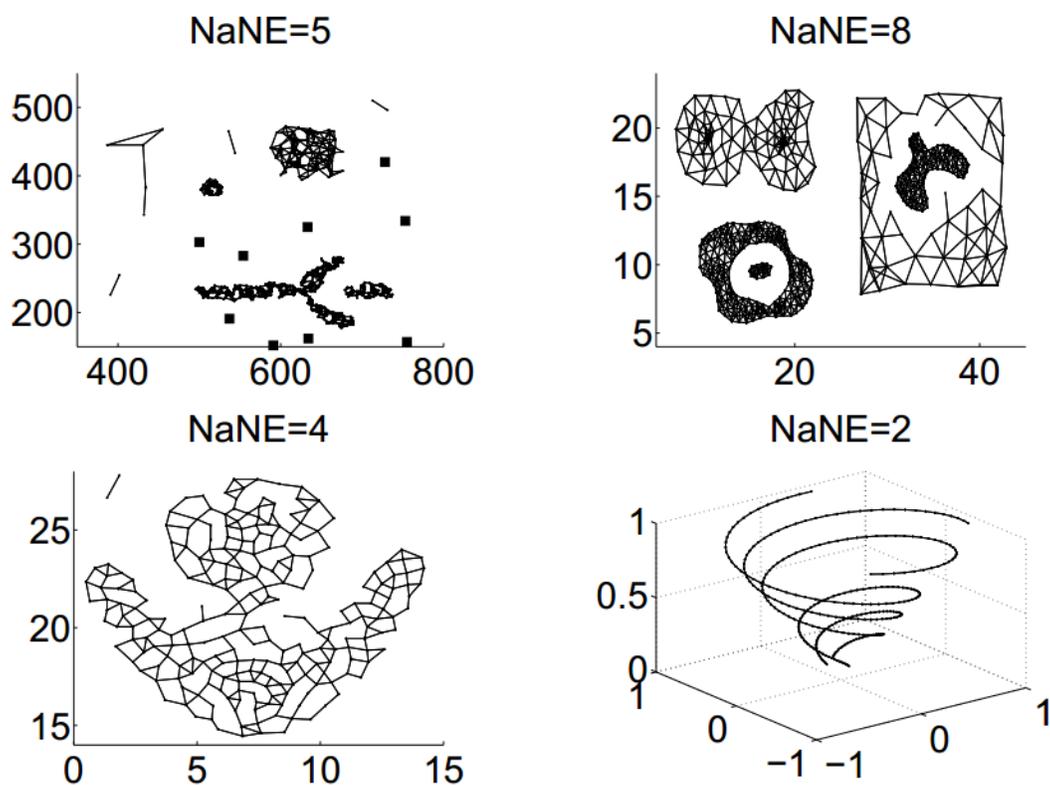

Figure 2: Natural neighbor graphs and natural neighbor eigenvalues in four different artificial data sets.

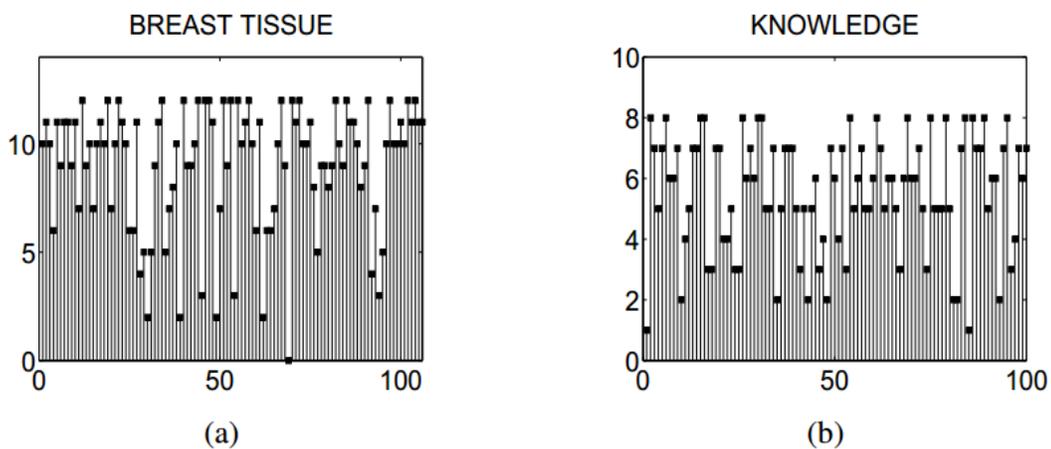

Figure 3: Dynamic neighborhood value of natural neighbor method. In order to show the results more clearly, here only shows the first 100 points in Fig.b

*3.2. Construct an efficient classification algorithm using ENN and NaN methods*

Our goal is to construct an efficient classification algorithm for nearest neighbor classification, with the help of ENN and NaN methods. This ENaN method performs two stages: training stage and testing stage, and these steps are explained below.

In training stage, the algorithm deals with training data, and thus NaNE of NaN method is used to be the parameter $k$ in the generalized class-wise statistic calculation step. The weighted KNN graphs can then be used to calculate distance and natural neighbor efficiently for a given test sample, so during the progress of generalized class-wise statistic calculation step, the algorithm stores the weighted KNN graphs and transfer it to the next stage. The following is the description of ENaN classification algorithm in training stage.

---

**Algorithm 1** ENaN classification algorithm: training stage

**Require:** training data set $X_{training}$;

**Ensure:** weighted KNN graph, $G = <V, E>$; generalized class-wise statistic for each class, $T = \{T_1, T_2, \ldots, T_c\}$;

1: Create a $k$-d tree *Tree* from training data set $X_{training}$.
2: Calculate *NaNE* of training data set $X_{training}$ by using $k$-d tree *Tree*.
3: Let *NaNE* be the parameter $k$ to build weighted KNN graph $G$ by using $k$-d tree *Tree*
4: **For all** class of training data set $X_{training}$ **do**
5:   Calculate the generalized class-wise statistic $T_i$ by using weighted KNN graph $G$
6: **end for**
7: Return: $G = <V, E>$, $T = \{T_1, T_2, \ldots, T_c\}$

---

The time complexity of algorithm 1 is $O(m * \log m)$, m is the size of training data set. Algorithm 1 firstly creates a $k$-d tree of the data set, and the time complexity of this step is $O(m * \log m)$. After that, the complexity of *NaNE* calculation is $O(NaNE * m * \log m)$. The value range of *NaNE* must $2 \leq NaNE < m$, it is generally 6 or 7, and for high dimensional or irregular data sets, the *NaNE* will be more than 20 but less than 30. So in the *NaNE* calculation step, the complexity is $O(m * \log m)$. For each class, to build weighted KNN graph at a computational complexity of $O(m * \log m)$. At last, we do generalized class-wise statistic computation with only computational complexity of $O(m)$.

In testing stage, the goal of the algorithm is to determine which class the sample in the testing data set belongs to. Taking into account the diversity of samples, here we use the number of every samples' natural neighbors as its neighborhood parameter to calculate the generalized class-wise statistic. The following is the description of ENaN classification algorithm in testing stage.

---

**Algorithm 2** ENaN classification algorithm: testing stage

**Require:** training data set $X_{training}$; testing data set $X_{testing}$; weighted KNN graph, $G =< V, E >$; generalized class-wise statistic for each class, $T = \{T_1, T_2, \ldots, T_c\}$;

**Ensure:** label set of testing data;

1: $\forall x_i \in X_{testing}$, $NaN\_Num(x_i)=0$

2: **for all** sample $x_i$ in testing data set $X_{testing}$ **do**

3:     Calculate the number of its natural neighbor $numNaN(x_i)$ in data set $X_{training}$ by using weighted KNN graph $G$

4:     **for all** class of training data set $X_{training}$ **do**

5:         Assume sample $x_i$ belongs to current class

6:         Calculate the generalized class-wise statistic $T_i$ by using weighted KNN graph $G$

7:     **end for**

8:     Calculate ENN classification prediction $f_{ENN}$

9:     Make classification decision of sample $x_i$

10: **end for**

11: Combine all samples' classification decisions as label set of testing data

12: Return: label set of testing data

---

The time complexity of algorithm 2 is $O(m * n)$, $m$ is the size of training data set, and $n$ is the size of testing data set. Because of the existence of weighted KNN graph, for any sample in the testing data set, it is only necessary to calculate the number of its natural neighbor in the training data set at a computational complexity of $O(m)$. Similarly, the time complexity of generalized class-wise statistic calculation step in this algorithm is $O(m)$.

4. **Experimental evaluation**

To demonstrate the superiority of the proposed algorithm over original ENN algorithm for classification, in this section we compare the original ENN algorithm with our ENaN method. The best $k$ in original ENN algorithm is searched in the range: *1*, *3*, *5*[20] and *sqrt(n)* [21] via 10-fold cross-validation on the training set. All algorithms and experiments run on Matlab R2014a.

All 15 real world data sets are obtained from the University of California, Irvine (UCI) machine learning repository [22]. Table 1 shows the properties of these data sets.

Table 1: Data size and dimensions of all UCI data sets

| Data set | size | Dimensions |
|---|---|---|
| DIABETES | 768 | 8 |
| GLASS | 214 | 9 |
| ECOLI | 336 | 7 |
| HABERMAN | 306 | 3 |
| SEGEMENT | 2310 | 19 |
| SEGEMENT_TRAIN | 210 | 13 |
| WINE | 178 | 13 |
| IRIS | 150 | 4 |
| SONAR | 208 | 60 |
| VEHICLE | 846 | 18 |
| CANCER | 569 | 30 |
| LIBRAS | 900 | 90 |
| LETTER | 2000 | 16 |
| PAGEBLOCKS | 5473 | 10 |
| KNOWLEDGE | 403 | 5 |

We evaluate their performance in term of classification accuracy. The results of all 15 UCI data sets are presented in Table 2.

Table 2: Classification accuracies of ENN algorithms over a range of *k* and ENaN algorithm with no neighborhood parameter

| Data sets | k=1 | k=3 | k=5 | k=sqrt | ENaN |
|---|---|---|---|---|---|
| diabetes | 100.00±0% | 100.00±0% | 100.00±0% | 100.00±0% | 100.00±0% |
| glass | 55.74±5.22% | 57.53±5.34% | 56.65±6.23% | 45.00±7.34% | 47.36±5.53% |
| ecoli | 73.77±3.38% | 74.09±4.64% | 77.34±4.25% | 79.77±2.88% | 79.79±2.94% |
| haberman | 69.81±1.38% | 60.68±2.15% | 68.22±0.73% | 73.76±0.80% | 71.82±0.79% |
| segment | 95.89±0.02% | 96.45±0.01% | 95.45±0.01% | 86.97±0.06% | 90.74±0.07% |
| segment_train | 66.67±6.10% | 65.24±6.45% | 66.67±7.26% | 65.24±8.27% | 70.48±5.78% |
| wine | 69.44±3.12% | 67.32±1.92% | 68.37±2.38% | 64.48±2.47% | 71.76±2.14% |
| iris | 96.00±0.32% | 95.33±0.40% | 94.00±0.64% | 96.00±0.32% | 95.33±0.30% |
| sonar | 43.21±3.07% | 45.10±3.16% | 40.69±4.07% | 43.64±2.36% | 39.74±3.60% |
| vehicle | 62.29±0.11% | 63.46±0.30% | 64.53±0.29% | 60.75±0.20% | 64.53±0.24% |
| cancer | 90.70±0.27% | 91.56±0.29% | 91.92±0.23% | 92.80±0.35% | 92.80±0.48% |
| libras | 33.33±0% | 33.11±0% | 30.44±0.10% | 22.22±0.75% | 31.33±0.06% |
| letter | 29.98±2.01% | 30.31±2.06% | 30.20±2.05% | 22.05±0.85% | 28.70±1.97% |
| pageblocks | 79.79±7.49% | 73.76±6.62% | 78.91±7.46% | 88.41±9.68% | 87.17±9.33% |
| knowledge | 83.13±0.23% | 85.84±0.11% | 87.85±0.17% | 87.33±0.15% | 89.58±0.34% |
| OVERALL | 69.98±2.18% | 69.32±2.23% | 70.08±2.38% | 68.56±2.43% | 70.74±2.24% |

From Table 2, our algorithm wins in 5 out of 15 cases over all other choices, and it is very close to the best one in most situations. Overall, ENaN algorithm achieves the highest accuracy and stability. Recall that, the proposed algorithm is a completely neighborhood parameter free algorithm, which means that we no longer need to choose the adaptive k for each data set, this advantage is even more important than the classification accuracy.

This result demonstrates the universality of ENaN algorithm, and the characteristics of this neighborhood parameter free method exists in most nearest neighbor based algorithm. Therefore, we have the chance to innovate and solve more problems in clustering and outlier detection with the proposed NaN based method, or more NaN based method.

**5. Conclusion**

This paper proposes a novel parameter free classification method based on extended nearest neighbor method, the parameter k choosing problems in training stage and testing stage are perfectly solved by natural neighbor method in uncorrelated ways. Experimental results show that our classification result intuitively reflects the characteristics of data sets, and compared to the original ENN algorithm in different parameter k, our algorithm increases the accuracy of classification result as well as its adaption to different kind of data sets, and avoid the neighborhood choosing problem.

Acknowledgements

This work was supported by the National Nature Science Foundation of China (No. 61272194, No. 61502060 and No. 61073058)